\def\year{2017}\relax
\documentclass[letterpaper]{article} 
\usepackage{aaai18}  
\usepackage{times}  
\usepackage{helvet}  
\usepackage{courier}  
\usepackage{url}  
\usepackage{graphicx}  
\usepackage[cmex10]{amsmath}
\usepackage{subfigure}
\usepackage[ruled,linesnumbered]{algorithm2e}
\usepackage{multirow}
\graphicspath{ {images/pdf} }

\begin{document}
%
\title{Towards Imperceptible and Robust Adversarial Example Attacks against Neural Networks}

\author{Bo Luo, Yannan Liu, Lingxiao Wei, Qiang Xu\\
Department of Computer Science \& Engineering\\
The Chinese University of Hong Kong\\
\{boluo,ynliu,lxwei,qxu\}@cse.cuhk.edu.hk
}
%

\maketitle

\begin{abstract}
Machine learning systems based on deep neural networks, being able to produce state-of-the-art results on various perception tasks, have gained mainstream adoption in many applications. However, they are shown to be vulnerable to adversarial example attack, which generates malicious output by adding slight perturbations to the input. Previous adversarial example crafting methods, however, use simple metrics to evaluate the distances between the original examples and the adversarial ones, which could be easily detected by human eyes. In addition, these attacks are often not robust due to the inevitable noises and deviation in the physical world. In this work, we present a new adversarial example attack crafting method, which takes the human perceptual system into consideration and maximizes the noise tolerance of the crafted adversarial example. Experimental results demonstrate the efficacy of the proposed technique.

\end{abstract}


%

\section{Introduction}
With the increasing popularity of deep learning technology, its security problems have attracted a significant amount of attention from both academia and industry.
Among all security problems, the most severe one is adversarial example attack first proposed in~\cite{szegedy2013intriguing}. It attempts to modify the legal input with slight perturbations that largely changes the output given by neural networks. This kind of attack is really a threat for security-sensitive systems, such as self-driving systems, disease diagnose systems and malicious email filters~\cite{bojarski2016end,amato2013artificial,clark2003neural}. For example, in self-driving systems, an adversarial example attack can change a stop-road sign to a turn-left signal. Then the car will make a wrong decision and cause a serious traffic accident. 

Several adversarial example attacks against neural networks are proposed in the literature~\cite{szegedy2013intriguing,papernot2016limitations,goodfellow2014explaining}. They target to misclassify the output to a specific class by adding minimum perturbations. Although they demonstrate effective attacks against neural networks, several problems remained to be solved.
Firstly, adversarial examples generated by them are not sufficiently imperceptible. They all use distance metrics of $L_p$ norms ($L_0$, $L_2$ and $L_\infty$ norms) to evaluate the similarity between the original samples and the crafted adversarial ones. 
These metrics are objective, which treat perturbations of different pixels in an image equally important for human eyes. However, according to~\cite{legge1980contrast}, people are more sensitive to perturbations of pixels in low variance regions. For example, perturbations in the uniform background are easier to be detected than those in image regions with mussy objects. Without considering the human perceptual system, highly sensitive pixels may be changed too much, which increases the probability to be detected. 
Secondly, previous attacks are not robust enough. The success attack rate drops largely in the physical world due to the noises and deviation inevitably generated. For instance, the adversarial examples may be compressed or suffer from noises during transmission. Thus, the adversarial example which attacks successfully in the experimental condition may fail in the complex physical world. Recently, some research efforts have been dedicated to robust attacks for certain situations, such as face recognition~\cite{sharif2016accessorize} and road sign recognition~\cite{evtimov2017robust}. However, they are rather application-specific and cannot be generalized for other applications.

To solve the above problems, in this paper, we propose a new method to craft imperceptible and robust adversarial examples against neural networks. We first introduce a new distance metric considering sensitivity of the human perceptual system to different pixels. This metric guides us with how many perturbations can be added without being detected. 
Then we optimize to maximize the noise tolerance of adversarial examples to improve the success attack rate in the physical world. Specifically, we target to increase the gap between the target class probability and the max probability of other classes, which is generally applicable for a large amount of applications based on neural networks. By introducing a new quantity to evaluate the effects of perturbations added to each pixel, we present a greedy algorithm to find which pixels to perturb and what magnitude to add effectively and efficiently. Our optimization method can generate adversarial examples with both high imperceptibility and high robustness, as demonstrated in the experimental results.

\textbf{}

\textbf{}

\section{Related Work}
\label{related work}


Szegedy \emph{et al.} first proposed adversarial example attack against neural networks. It minimizes the distances evaluated with $L_2$-norm between the original examples and the adversarial ones under the constraint that the adversarial attack is successful.
FGSM~\cite{goodfellow2014explaining} performs the attack by first calculating the gradients of the loss function to search which directions to change for each pixels. Then it modifies all pixels simultaneously under the $L_\infty$ constraint.
Recently, JSMA~\cite{papernot2016limitations} builds a saliency map to model the impact each pixel has on the resulting classification. It then optimizes with the $L_0$ distance, where it picks the most important pixel based on the saliency map and modify the pixel to increase the target class probability in each iteration. However, these attack methods all use simple distance metrics ($L_p$-norms) to evaluate the similarity between the adversarial example and the original one without considering the human perceptual system. 


There are some research efforts about robust adversarial example attacks in the literature. The authors in~\cite{kurakin2016adversarial} first discussed the idea when they found some adversarial examples survived in the physical world. However, they did not present a solution to improve the success attack rate. Recently, two papers studied adversarial examples for certain applications in the physical world. \cite{sharif2016accessorize} proposed a physical realizable adversarial example attack against the face recognition system through wearing malicious eye-glasses. \cite{evtimov2017robust} discussed a practical attack on the road sign recognition in self-driving systems. They generate adversarial road signs which can successfully deceive the recognizer in various directions and angles. However, these two methods are rather application-specific and are not generally applicable.

Apart from adversarial example attacks against neural networks in computer vision systems, many other machine learning applications are suffering from adversarial example attacks. \cite{carlini2016hidden} proposed an attack against the speech recognition system, where they show how to craft sounds that are difficult for human to understand, but can be interpreted to specific commands such as ``Call 911'' and ``Turn on airplane mode''.  In~\cite{grosse2016adversarial}, they introduce adversarial example attacks against malware detecting systems. In these attacks, they disguise a malware into a benign one and successfully fool the detector.
\section{Adversarial Example Attacks}
Adversarial example attacks target to change the output of machine learning systems by adding slight perturbations to the input. In the literature, there are two categories of adversarial example attacks: target attack and un-target attack. For the target attack, it attempts to misclassify a sample to a specific class, while un-target attack only tries to misclassify the input. As a result, the target attack is more difficult than the un-target one. In this paper, we focus on the target adversarial example attack.

Generally speaking, adversarial example attacks should not only fool machine learning systems but also consider two important factors: imperceptibility and robustness.

\textbf{Imperceptibility: }
The imperceptibility of an adversarial example means that it should look similar to the original one in order not to be detected by human eyes. So in the attack, it is important to use an appropriate distance metric to evaluate the similarity between an adversarial example and the original one. A good distance metric should clearly reflect the characteristic of the human perceptual system. Otherwise, the imperceptibility of adversarial examples would not be ensured. 



\textbf{Robustness: }
Adversarial examples are firstly crafted by the attackers and then transmitted to the machine learning systems. They may fail to attack after the transmissions with inevitable noises or deviation. The robustness of adversarial examples reflects its ability to stay misclassified to the target class after suffering transformations in the physical world. The definition is as following:
\begin{equation}
\begin{array}{c}
F(X^*)=T, \\
\\
F(Tran(X^*))=T,
\end{array}
\end{equation}
where $X^*$ is the adversarial example crafted by the attacker, $T$ is the target class specified, and $Tran(*)$ is the transformation suffered in the physical world. Previous methods do not consider the robustness and thus adversarial examples crafted by them may be largely destroyed and fail to attack in the physical world.

In this paper, we propose a new crafting method to generate adversarial examples with both high imperceptibility and high robustness, as detailed in the following sections.

\section{The Proposed Method}
In this section, we first present a new distance metric considering the effects of different pixels on human eyes, then we propose to maximize the probability gap between the target class and left classes for increasing the noise tolerance of adversarial examples. As last, an efficient greedy algorithm is introduced to enhance robustness and ensure imperceptibility in adversarial attacks.

\subsection{Imperceptibility of Adversarial Example Attacks}

\begin{figure}[h]
\centering
\includegraphics[width=1.0\linewidth]{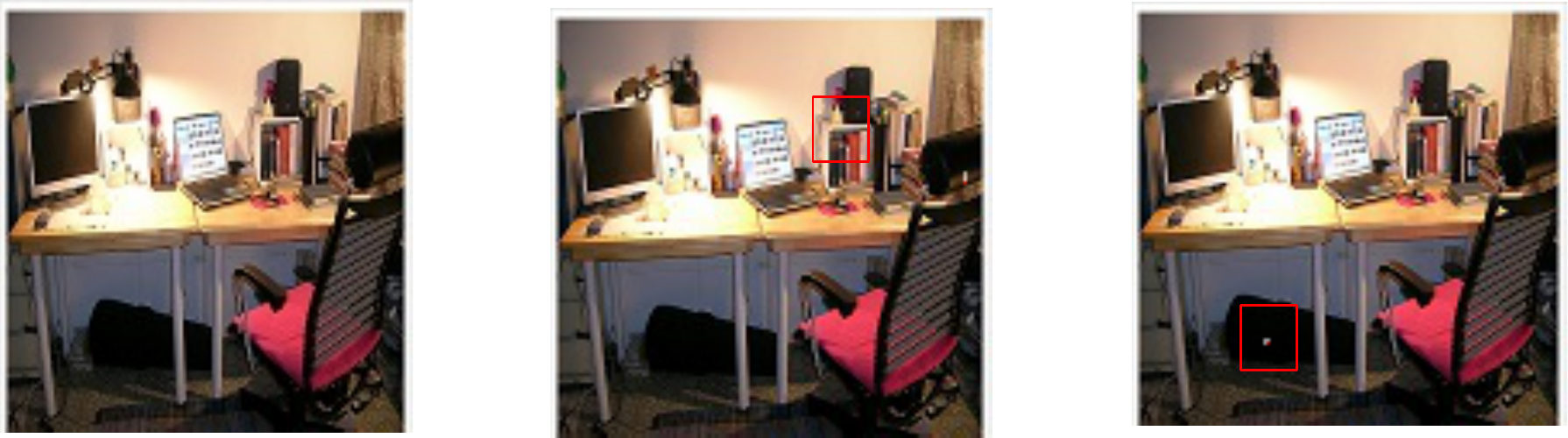}
\caption{Perturbations added in different pixels raise varying human perceptual attentions. The red line box marks the perturbation in each perturbed image.}
\label{fig:sensitivity}
\end{figure}

According to contrast masking theory in image processing~\cite{legge1980contrast,lin2005visual,liu2010just}, human eyes are more sensitive to perturbations on pixels in low variance regions than those in high variance regions. 
For example, in Figure~\ref{fig:sensitivity}, the left image is the original sample. The middle and right images are perturbed with the same magnitude on 10 pixels but at different positions. The positions of perturbations on the middle image are all at high variance region (the disorder desk) while the right image is perturbed at the low variance region (the black bag on the floor). We can see that it is hardly to detect the perturbations on the middle image, however, people with normal visual capability can notice the perturbations on the black bag on the floor.

Therefore, to make adversarial examples imperceptible, we should perturb pixels at high variance zones rather than low variance ones. In this paper, we compute the variance of a pixel $x_i$ based on the standard deviation $SD(x_i)$ among an $n\times n$ region as shown in Equation~\ref{eq:standard_deviation}, where $S_i$ is the set consisting of pixels in the $n\times n$ region, $\mu$ is the average value of pixels in the region. Specifically, for $n=3$, the variance is calculated as the standard deviation of the pixel and its 8 neighbors. 
\begin{equation}\label{eq:standard_deviation}
SD(x_i)= \sqrt\frac{\sum\limits_{x_k \in S_i}(x_k - \mu)^2}{n^2}.
\end{equation}


Accordingly, we introduce \emph{perturbation sensitivity} to measure how much ``attention'' will be drawn by adding per ``unit'' perturbation on a pixel. It is defined as follows:
\begin{equation}
Sen(x_i)= 1/ SD(x_i).
\end{equation}
When the pixel has low variance, the perturbation sensitivity is high. Therefore, adding perturbations on this pixel is easily detected by humans.
%

To evaluate the human perceptual effect of a perturbation added to a pixel, we can multiply the magnitude of the perturbation by its sensitivity. When crafting an adversarial example, we usually perturb more than one pixel. As a result, we sum up all the effects of perturbations and use it as the distance between the original example and the adversarial one, as shown in the following equation:


\begin{equation}
D(X^*, X)= \sum_{i=1}^N \delta_i *Sen(x_i),
\end{equation}
where $D(X^*, X)$ denotes the distance between the adversarial example $X^*$ and the original one $X$. $\delta_i$ is the perturbation added to the pixel $x_i$ and $N$ is the total number of pixels. When generating adversarial examples, we can add constraints on the distance so that the perturbations would not be detected.

\subsection{Robustness of Adversarial Example Attacks}

Another limitation of existing methods for adversarial example attack is that they have very low success rates in the physical world due to deviation caused by regular transformations of images such as compressing, resizing and smoothing. 
The challenge of the problem is that transformations in the physical world are usually uncertain and hard to model, and thus we cannot enhance robustness of attacks for specific situations. In this paper, we give a general solution for robust attacks by maximizing noise tolerance of adversarial examples. The noise tolerance reflects the amount of noises that adversarial examples can tolerate with the misclassified target label unchanged.

Classifiers based on neural networks output the probabilities for all classes, and select the highest one as the result label for the given input. The probability for one class denotes the confidence of classifying the input to this category. Previous adversarial example attacks only maximizes the probability of the target class, however, we find for robust attack, it is necessary to reduce the max probability among other classes as well. Naturally, we dedicate to maximize the gap between the probability of the target class and the max probability of all other classes. It can be formulated as follows:
\begin{equation}
Gap(X^*)= P_t - max(P_i)\qquad (i\neq t),
\end{equation}
where $P_t$ denotes the target class probability and $P_i$ refers to probabilities of other classes. Intuitively, the higher the probability gap, the more robust adversarial example attacks.

\begin{figure}[h]
\centering
\includegraphics[scale=0.82]{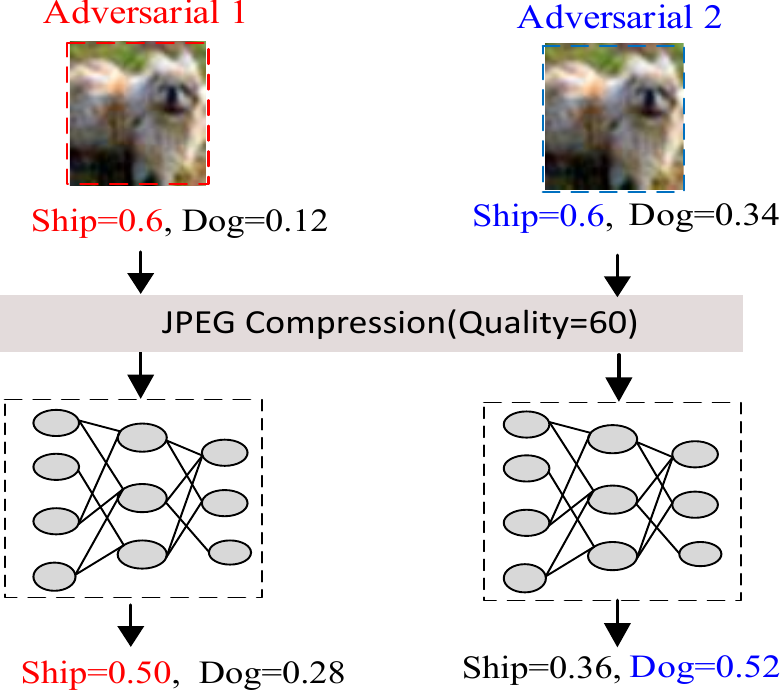}
\caption{Adversarial example with higher probability gap is more robust when suffering a same image transformation. }
\label{fig:gap}
\end{figure}

Figure~\ref{fig:gap} is a simple example to illustrate this idea. In this figure, you can see there are two adversarial examples against the same original sample. They are all misclassified as ship with 0.6 probability, but with different probability gap. Adversarial example 1 has a higher probability gap than adversarial example 2. Now, after suffering JPEG compression (quality is 60), adversarial example 1 still is classified as ship with 0.5 probability. While adversarial example 2 is classified as dog with 0.52 probability and the probability of ship now decreases to 0.36. Only the first two classes with the highest probability are listed in the Figure.

\subsection{Imperceptible and Robust Attacks}

In this work, we target to achieve both imperceptible and robust adversarial example attacks against neural networks. In order to obtain imperceptibility, the distance $D(X^*, X)$ between the original example $X$ and the adversarial one $X^*$ should be constrained. Then we can increase the robustness of adversarial example attacks whenever possible under this constraint. Overall, we formulate the problem as follows:
\begin{equation}
\begin{array}{c}
\underset{X^*}{\operatorname{argmax}} \;\; Gap(X^*) \\
\\
s.t.\;\;D(X^*,X)\leq D_{max}\\
\end{array}
\end{equation}
where $D_{max}$ is the largest distance allowed in order not to be detected by human eyes. In practice, users need to determine this value based on the input images.

This problem is difficult to solve as the objective function (probability gap) is not differentiable including a max function (the max probability of left classes). Thus the traditional gradient descent method is not applicable. Moreover, in each iteration, it is time consuming to determine which pixels to perturb and what magnitudes to add. To deal with these problems, we first smooth the objective function to make it differentiable and then propose an efficient greedy algorithm to simplify computations in each iteration, as detailed in the following subsections.



\subsubsection{Smoothing the Objective Function}

We smooth the $max$ function to the differentiable one based on the following equation:
\begin{equation}
    \begin{array}{c}

max(x,y) \approx \log(e^{kx}+e^{ky})/k. \\
   \end{array}
\end{equation}

It achieves the approximation by amplifying the difference between $kx$ and $ky$ using the exponential function. When $kx$ is quite bigger than $ky$, $e^{kx}$ will be much larger than $e^{ky}$. Then $e^{kx} + e^{ky}$ will approximately equal to $e^{kx}$ and $log(e^{kx})/k$ is essentially equal to $x$. When $x$ and $y$ are not significantly different, $k$ is used to improve the accuracy of approximation. Given an example with $x = 0.2, y=0.1$, if $k=1$, then $\log(e^{kx}+e^{ky})/k \approx 0.84$. However, it approximates to $0.2000005$ when $k=100$. We can make the approximation as close to the max function as we want by setting large enough $k$. 


Now the objective function is transformed in the following format and can be differentiated for further optimization:
\begin{equation}\label{eq:objective}
    \begin{array}{c}
Gap(X^*)\approx P_t -\log(\sum{e^{kP_i}})/k \qquad i\neq t
    \end{array}
\end{equation}

\subsubsection{A Greedy Algorithm for Optimization}
After smoothing the objective function, we can solve the problem using the traditional gradient descent method. However, in each iteration, we have to choose which pixels to modify and what magnitudes to add. Even though we assume each pixel is perturbed with the same magnitude, the time taken for solving the problem is still prohibitively long. For example, if each image contains 100 pixels and we choose to perturb 10 pixels at each iteration, then we have to search $100 \choose 10$ times to find the best 10 pixels to modify. 

Considering that we have to choose pixels with less perturbation sensitivity to human eyes and at the same time increase the objective function in Equation~\ref{eq:objective}, we define a  new quantity called \emph{perturbation priority} to estimate the effect of perturbing a pixel:

\begin{equation}\label{eq:attack_intensity}
PerturbPriority(x_i)= \frac{\bigtriangledown_{x_i} Gap{(X^*)}}{Sen(x_i)},
\end{equation}
where $\bigtriangledown_{x_i} Gap{(X^*)}$ is the gradient of the probability gap for pixel $x_i$. Perturbation priority indicates how much probability gap increased by adding one ``unit'' of perturbation to the current pixel $x_i$, and therefore it reflects the priority of pixels to perturb in the adversarial example generating process. 
\begin{figure*}[ht]
\centering
\includegraphics[scale=0.42]{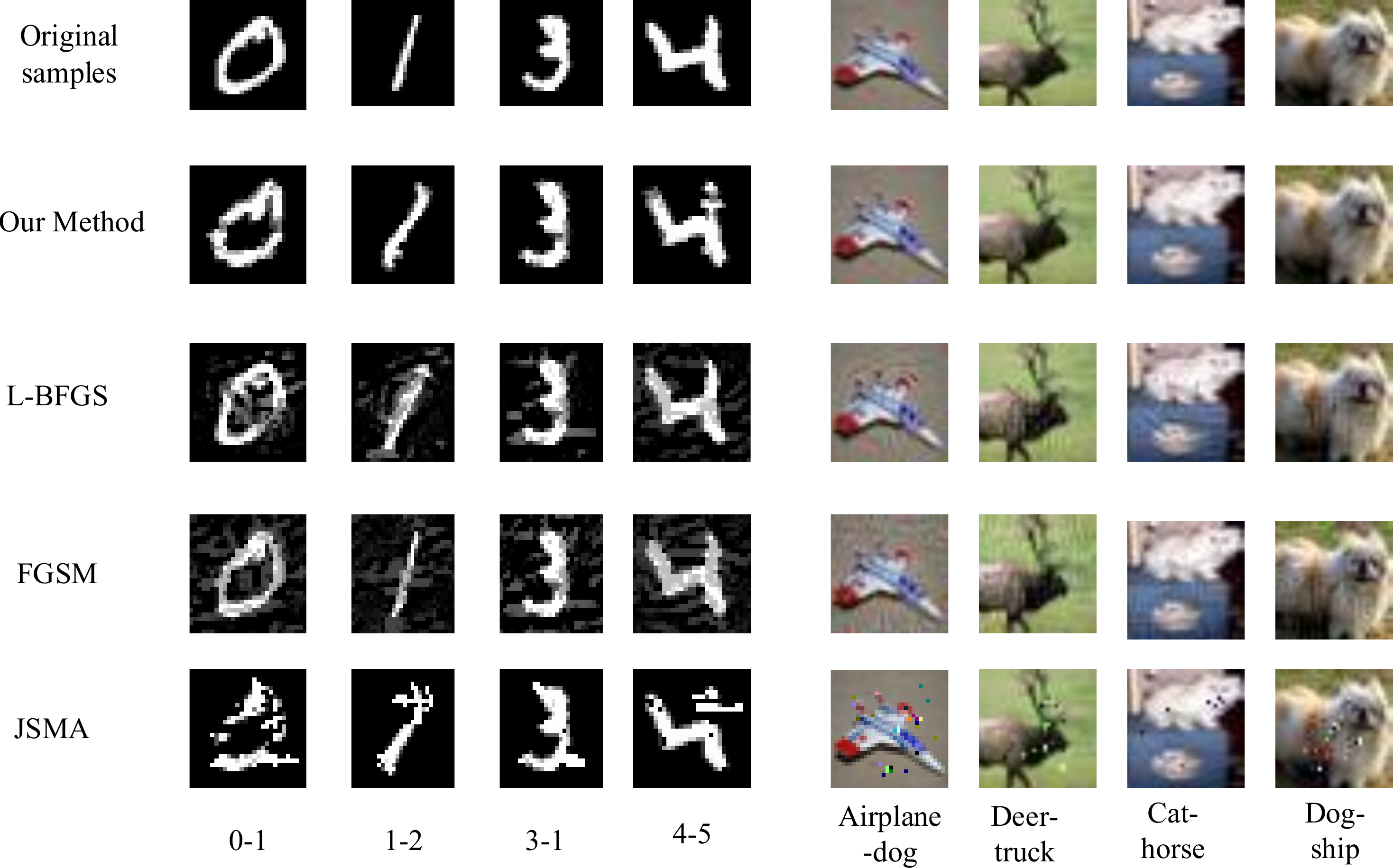}
\caption{Adversarial examples generated by different crafting methods in MNIST and CIFAR10. Adversarial examples in the second row crafted by our method are much more imperceptible than others from the following rows. While JSMA method in the last row performs the worst.}
\label{fig:image}
\end{figure*}

\begin{algorithm}[h]\label{alg:process}
\KwIn{The legitimate sample $X$, the max allowed human perceptual distance $D_{max}$, the number of pixels perturbed in each iteration $m$ and the perturbation magnitude $\delta$.}
\KwOut{Adversarial example $X^*$.}
\While{$D(X,X^*)< D_{max}$}{
   $PerturbPriority$ $\leftarrow$ Calculate perturbation priority for each pixel; \\
   $SortedPerturbPriority$ $\leftarrow$ Sort perturbation priority in $PerturbPriority$;\\

    $SelectedPixels$ $\leftarrow$ Choose $m$ pixels with largest perturbation priority; \\

    $X^*$ $\leftarrow$ Perturb selected pixels with magnitude $\delta$;\\

    $D(X^*,X)= \sum_i^N \Delta_i * Sen(x_i)$; \\

    $X=X^*$.
}

\caption{The proposed algorithm to generate adversarial examples.}
\end{algorithm}

Based on the perturbation priority, we propose a greedy algorithm to efficiently achieve imperceptible and robust adversarial example attacks. The detailed crafting process is shown in Algorithm~\ref{alg:process}, in which it first calculates each pixels' perturbation priority based on the gradients of the probability gap and perturbation sensitivity in line 2. Then we sort pixels according to perturbation priority in line 3. Next, we perturb the first $m$ pixels with a small magnitude $\delta$ and calculate the human perceptual distance of the updated adversarial examples in line 4-6, where $\Delta_i$ is the total perturbations added to $x_i$. At last, in line 7 the original example $X$ is updated. The whole process is repeated until the constraint on $D(X^*, X)$ is violated.


\section{Experimental Evaluations}
\textbf{}

\textbf{Dataset.} All the experiments are performed on MNIST and CIFAR10 datasets. The MNIST dataset includes 70000 gray scale hand-written digit images with the size of 28*28. The classification goal is to map the images to the corresponding digits from 0 to 9. The CIFAR10 dataset contains 6000 color images. Each image has the size of 32*32*3. There are 10 classes in the dataset, which are airplane, automobile, bird, cat, deer, dog, frog, horse, ship and truck. The intensity values of pixels in all these images are scaled to a real number in $[0,1]$.

\textbf{DNN Model.} For each dataset, we trained a model. The architectures of these two models are detailed in Table~\ref{tab:model}. They are all 8 layers DNN with ReLU as the activation function. The MNIST and CIFAR10 model achieve 99.18\% and 84.21\% classification rate respectively.

\begin{table}[h]
\centering
\caption {Model architectures.} \label{tab:model}
\begin{center}
\begin{tabular}{p{3.4cm} p{1.5cm} p{1.5cm}}
 \hline

  Layer & MNIST &CIFAR \\
  \hline
  Input layer&28, 28&32, 32, 3\\
  Convolution layer 1& 3, 3, 32&3, 3, 64 \\
  Convolution layer 2& 3, 3, 32&3, 3, 64 \\
  Max pooling layer 1 & 2, 2&2, 2\\
  Convolution layer 3 & 3, 3, 64&3, 3, 128\\
  Convolution layer 4 & 3, 3, 64&3, 3, 128\\
  Max pooling layer 2 & 2, 2&2, 2\\
  Fully connected layer 1& 128 & 512\\
  Fully connected layer 2& 10 & 10\\
  \hline
  \multicolumn{3}{c}{Softmax}\\
 \hline
\end{tabular}
\end{center}
\end{table}

\textbf{Baselines.}
The baselines used in these experiments are three widely-used adversarial example attacks, Jacobian-based Saliency Map Approach (JSMA)~\cite{papernot2016limitations}, iterative Fast Gradient Sign Method (FGSM)~\cite{goodfellow2014explaining} and box-constrained L-BFGS method~\cite{szegedy2013intriguing}. For detailed experimental setups of these methods please refer to the original papers. 
In our method, we select 20 pixels to add perturbations with a magnitude of 0.01 in each iteration.

\subsection{Evaluate Imperceptibility}

In this experiment, we evaluate the imperceptibility of adversarial examples crafted by our method and the baseline methods.
We perform adversarial example attacks against the testing set (10000 test images) in MNIST and CIFAR10 respectively. These adversarial examples are just successfully misclassified to the target classes which were randomly assigned. That is to say, we stop adding perturbations once the target class attack is successful. Then for each attack method, we get two groups of adversarial examples for the two datasets.

\textbf{Evaluate with Human Perception:}
We present several groups of images in Figure~\ref{fig:image}. Each group images include an original sample and its corresponding adversarial examples crafted by different attack methods.
The left four columns are from MNIST and the right four columns are from CIFAR10. Images in the first row are original samples in the testing set. The second row are adversarial examples crafted by our method. The following rows are adversarial examples from L-BFGS, FGSM and JSMA attack methods, respectively. The target adversarial classes are listed at the bottom.
From Figure~\ref{fig:image}, we can see that adversarial examples crafted by our method in the second row are the most imperceptible, which nearly look the same as the original one. While JSMA method in the last row performs the worst and the perturbed pixels are easily detected by human eyes. The reason is that JSMA method perturbs pixels to the maximum value without considering pixels' human perceptual sensitivity. As a result, the perturbed pixels may be in the high sensitive region and thus raise human attentions. For L-BFGS and FGSM methods, they perform better than JSMA. This is because these two methods use $L_2$ norm and $L_\infty$ norm which tend to perturb more pixels with smaller perturbations. Although the pixels perturbed may be in the sensitive region, they raise relatively low attentions with smaller perturbations.
These experimental results show that considering human perceptual system, our method can generate much more imperceptible adversarial examples comparing with baseline methods.

\textbf{}

\begin{figure}[h]
\centering
\includegraphics[scale=0.72]{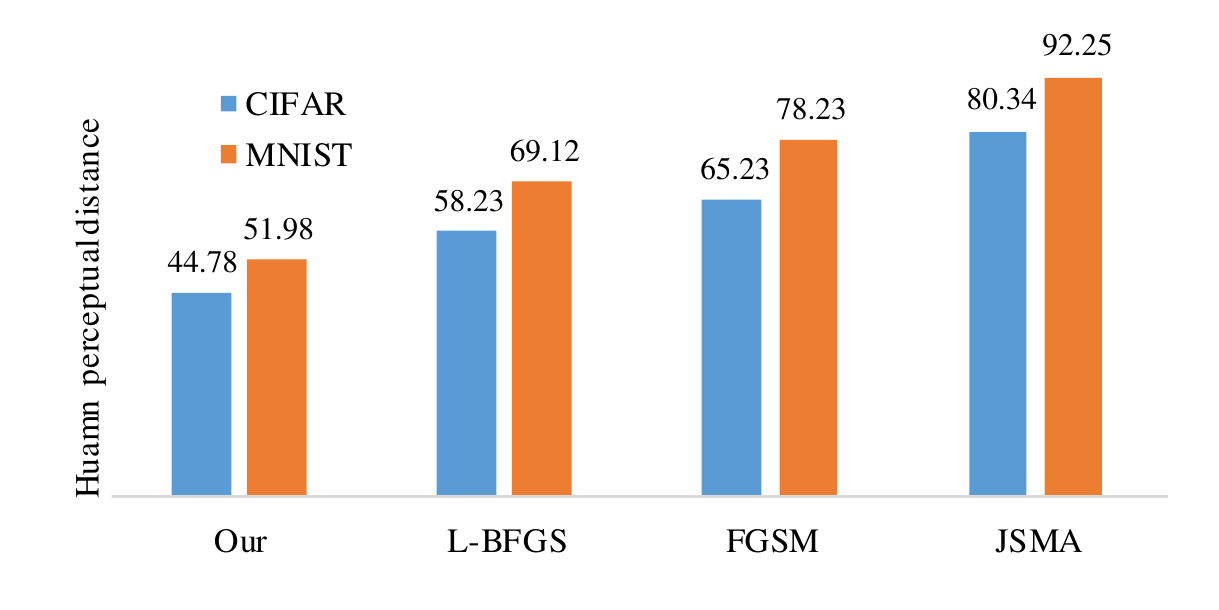}
\caption{Human perceptual distances of adversarial exampels crafted by varying adversarial methods in MNIST and CIFAR10. Adversarial examples crafted by our method has the smallest human perceptual distances.}
\label{fig:human_distance}
\end{figure}

\textbf{Evaluate Distance Metric:}
To evaluate the effectiveness of our human perceptual distance metric, we calculate the distances between the adversarial examples and original samples. The results are listed in Figure~\ref{fig:human_distance}.

\begin{figure*}[htpb]
\centering
\includegraphics[scale=0.8]{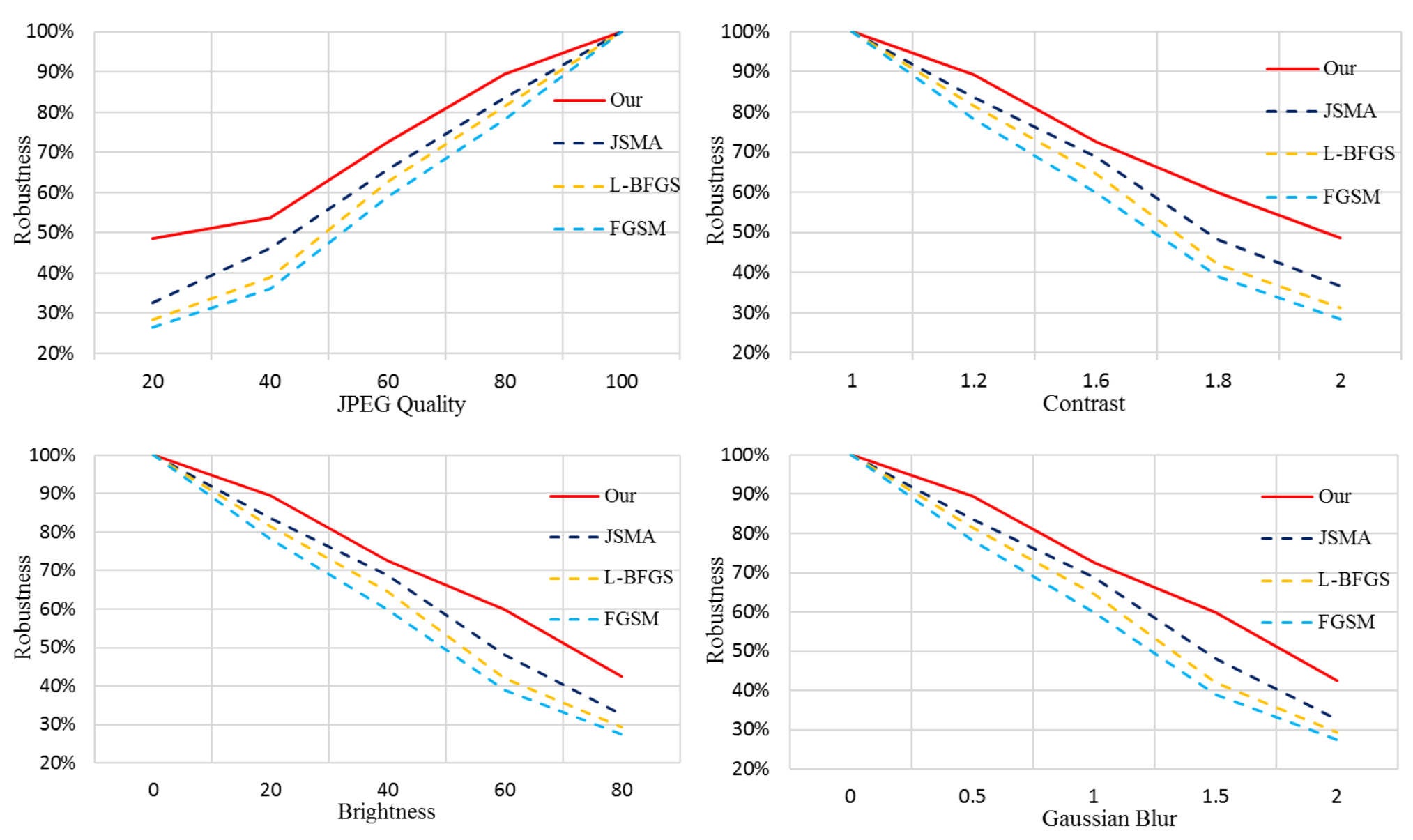}
\caption{Comparison of robustness for various adversarial methods for image transformations.}
\label{fig:transformations}
\end{figure*}

We can see that the distances of adversarial examples crafted by our method are the smallest (44.78 for MNIST and 51.98 for CIFAR10), while the distances of JSMA method are the largest (80.34 for MNIST and 92.25 for CIFAR10). This coincides with the results in the previous human perception experiments. Therefore, we can believe the distance metric proposed in this work can appropriately reflect the similarity between original samples and adversarial examples.
%

\textbf{Discussions:}
From the above results, we know that human perceptual distances in MNIST are larger than those in CIFAR10, so adversarial example attacks against MNIST dataset are more difficult than CIFAR10 dataset. We analyze the reasons from two aspects. One is that the images in MNIST have a large uniform background, while for CIFAR10, the backgrounds of natural images only occupy small regions. As a result, in MNIST images, pixels have a higher human perceptual sensitivity than those in CIFAR10.
The other reason is that the classification rate in MNIST dataset is about 15\% higher than CIFAR10. So the model trained in MNIST has higher confidence with the classifying results, thus it is more difficult to attack the model trained with MNIST dataset.

\subsection{Evaluate Robustness}
In this experiment, we evaluate the robustness of adversarial examples crafted with our method and the baseline methods.
We compare all the attack methods under the same max human perceptual distance, $D_{max} = 70$. This is determined empirically that the distance less than 70 would not raise much human attention.
In this part we only present the results for dataset CIFAR10, because the results for MNIST are quite similar.

\textbf{Robustness Definition:}
The robustness can be described as the fraction of adversarial examples which are still misclassified as the target class after the natural transformations. It is also called the success attack rate in the physical world. The definition is as follows:
\begin{equation}
R=\frac{\displaystyle\sum_{i=1}^{m}C(X_i,label_i)C(X^*_i,T_i)C(Tran(X^*_i),T_i)}
{\displaystyle\sum_{i=1}^{m}C(X_i,label_i)C(X^*_i,T_i)}
\end{equation}
where $m$ is the number of testing samples used to compute the robustness. $X_i$ is a test image and $label_i$ is the true label of this image, and $X^*_i$ is the adversarial example. $T_i$ is the target class for $X_i$ sample assigned by the attacker. The function $Tran(*)$ is an image transformation operation in the physical world. We study several transformations in this experiment, including adding gaussian noises, JPEG compressing, image blurring, changing contrast and brightness. The function $C(X,label)$ is used to check whether the image was classified correctly or not:
\begin{equation*}
C(X,label)=
\begin{cases}
1,& \text{If image $X$ is classified as $label$;}\\
0,& \text{otherwise.}
\end{cases}
\end{equation*}



%
\textbf{Evaluate Robustness with Gaussian Noises and Transformations:}
We test the physical success rate using four image transformations: JPEG compressing, gaussian blurring, contrast and brightness adjusting. The experimental results are showed in Figure~\ref{fig:transformations}.
We also test the robustness with gaussian noises which have five intensities with standard deviation changed from 0.05 to 0.25 with a step of 0.05 (The gaussian mean is 0). The experimental results are listed in Table~\ref{tab:noises}.

\begin{table}[htpb]
\centering
\caption {Comparison of robustness for various adversarial methods adding gaussian noises.} \label{tab:noises}
\begin{center}
\begin{tabular}{p{1.6cm}||p{1.2cm} p{1.2cm}p{1.3cm}p{1.2cm}}
 \hline
 Noises& Our&JSMA&L-BFGS&FGSM\\[0.16cm]
 \hline
 \\
 Std=0.05      &98.5\% &98.25\% &86.8\% &82.5\%\\[0.16cm]
 Std=0.1      &94.0\% &88.5\%  &82.0\% &79.5\%\\[0.16cm]
 Std=0.15     &77.8\%   &68.8\%  &62.6\% &64\%\\[0.16cm]
 Std=0.2      &68.5\% &55.12\% &50.8\% &42.5\%\\[0.16cm]
 Std=0.25       &62\%   &33.2\%  &28.6\% &21.5\%\\[0.16cm]
 \hline
\end{tabular}
\end{center}
\end{table}

It is clearly shown that our method performs the best among all the four transformations.
For example, in JPEG compressing, our method performs 76\% success attack rate while for FGSM method, the success rate is just 52.3\%.
Experimental results in adding gaussian noises show that our method also achieves higher robustness than other ones. Moreover, the benefit is more obvious with stronger noises. For example, in the fifth intensity with standard deviation (0.25), our method achieves 62\% success rate while the average success rate of the baseline methods is just about 26\%.

Apart from these observations, there are other observations drawn based on the results. Firstly, we can see that JSMA method performs the second best in these experiments though it achieves the worst results in previous human perception experiments. It can be explained that JSMA method perturbs less pixels with larger perturbations, and these large perturbations can tolerant more noises added on them. While for the FGSM method, it has the worst robustness in these experiments, because it tends to make small perturbations on the whole image. The effects of these small perturbations on pixels are more easily changed with noises. Our method, however, tries to maximize the noise tolerance and at the same time consider imperceptibility, therefore, it achieves great results for both imperceptibility and robustness. 
\section{Conclusions}
\label{conclusion}

Adversarial example attacks against neural networks have become one of the most severe security problems in artificial intelligence. Traditional adversarial example attacks do not consider human perceptual systems and thus are easily detected. Moreover, the success attack rate drops largely due to inevitable noises in the physical world. In this paper, we introduce a new adversarial example attack, which can achieve both high imperceptibility and robustness in the physical world. Specifically, we propose a new distance metric considering human perceptual systems. The metric evaluates the sensitivity of image pixel to human eyes, and thus it can guide us to add perturbations with less chances of being detected. To improve the successful attack rate in practice, we try to maximize the probability gap between the adversarial target class and other classes. A simple yet effective greedy algorithm is introduced to achieve the optimization goal under the constraint of not being detected. Experimental results show that adversarial examples achieved by our method is more imperceptible and robust than those produced by previous methods. 

\bibliographystyle{aaai}
\bibliography{ref}

\end{document}